\documentclass[11pt]{article}
\usepackage{eacl2017}
\usepackage{times}
\usepackage{url}
\usepackage{latexsym}
\usepackage{times}
\usepackage{graphicx,subfigure}
\usepackage{amssymb, amsmath}
\usepackage{soul}
\usepackage{multirow}

\eaclfinalcopy 

\title{A Joint Identification Approach for Argumentative Writing Revisions}

\author{Fan Zhang \\
  University of Pittsburgh  \\
   Pittsburgh, PA, 15260 \\
  {\tt zhangfan@cs.pitt.edu} \\\And
  Diane Litman \\
  University of Pittsburgh  \\
   Pittsburgh, PA, 15260 \\
  {\tt litman@cs.pitt.edu} \\}

\date{}

\begin{document}
\maketitle
\begin{abstract}
 Prior work on revision identification typically uses a pipeline method: revision extraction is first conducted to identify the locations of revisions and revision classification is then conducted on the identified revisions. Such a setting propagates the errors of the revision extraction step to the revision classification step. This paper proposes an approach that identifies the revision location and the revision type jointly to solve the issue of error propagation. It utilizes a sequence representation of revisions and conducts sequence labeling for revision identification. A mutation-based approach is utilized to update identification sequences. Results demonstrate that our proposed approach yields better performance on both revision location extraction and revision type classification compared to a pipeline baseline. 
\end{abstract}

\section{Introduction}
\label{intro}

Rewriting is considered as an important writing skill and researchers have demonstrated that experienced versus novice writers have different rewriting behaviors \cite{faigley1981analyzing}. Automatic revision identification allows the building of advanced writing tutoring systems that aim at improving students' writing skills \cite{roscoe2015automated,zhang-EtAl:2016:N16-3}. Revision identification typically involves two tasks: \textbf{revision extraction} where the locations of the revisions are identified and \textbf{revision classification} where the types of revisions are identified. Existing works typically follow a pipelined approach where the revision extraction step is first conducted (manually or automatically) and revision classification is conducted on the extracted revisions \cite{adler2011wikipedia,daxenberger2013automatically,bronner2012user,zhang-litman:2015:bea}. One problem of the pipelined approach is that the errors of the revision extraction step are propagated to the revision classification step. To solve this problem, an approach that can conduct revision extraction and revision classification at the same time is needed. 

In this paper we choose to conduct our study on argumentative revision detection \cite{zhang-litman:2015:bea,zhang-litman:2016:N16-1}. In \cite{zhang-litman:2015:bea}, revision locations are identified according to the result of sentence alignment and revision types are categorized to five categories according to their argumentation purpose: \textit{Claim}, \textit{Reasoning}, \textit{Evidence}, \textit{General} and \textit{Surface}\footnote{The categories are defined according to Toulmin's argumentation model ~\cite{toulmin2003uses}.}. Their experiment on pipeline revision identification demonstrates significant performance drop when compared to revision classification on gold-standard alignments. Table \ref{table:errorexample} demonstrates an example of error propagation in argumentative revision classification. According to human annotation, (D1-2) should be aligned to (D2-2), (D1-3) should be aligned to (D2-3). Based on alignment, their revision types should be \textit{Surface}\footnote{\textit{Surface} include changes such as spelling correction and sentence reorderings that do not change a paper's content.}. However, when the automatic sentence alignment misses the alignment, the revision classification step considers the sentences as deleted and categorizes them as \textit{Reasoning}. 

    We propose a sequence labeling-based joint identification approach by incorporating the output of both tasks into one sequence. The approach is designed based on two hypotheses. \textbf{First, the classification of a revision can be improved by considering its nearby revisions.} For example, a \textit{Claim} revision is likely to be followed by a \textit{Reasoning} revision\footnote{If you changed the thesis/claim of your essay, you have to change the way you reason for it.}. Zhang and Litman \shortcite{zhang-litman:2016:N16-1} used the types of revisions as labels and transformed the revision classification task to a sequence labeling problem. Their approach demonstrated significantly better performance than SVM-based classification approaches. In this paper, we extend their ideas by introducing \textbf{EditSequence} to also utilize alignment information for revision type prediction. An EditSequence describes a consecutive sequence of edits where not only the revision type but also the alignment information are incorporated into the labels of the edits. We hypothesize that adding alignment information can further improve revision type prediction. \textbf{Second, the alignment of sentences can be corrected according to the types of labeled revisions.} For example, the predicted types in Table \ref{table:errorexample} as a whole are rare, as there are 2 deleted \textit{Reasoning} sentences and 2 added \textit{Reasoning} sentences without any \textit{Claim} change. Such a sequence is likely to have a small likelihood in sequence labeling and thus a possible alignment error is detected. We introduce the idea of ``mutation'' from genetic algorithms to generate possible corrections of sentence alignments. The alignment of sentences after correction allows us to conduct a new round of revision type labeling. Our approach iteratively mutate and label sequences until we cannot find sequences with larger likelihood. Two approaches are utilized to generate seed sequences for mutation: (1) Direct transformation from predicted sentence alignment \cite{zhang-litman:2014:W14-18} (2) Automatic sequence generation using a Recurrent Neural Network (RNN). These settings together allow us to achieve better performance for both revision extraction and revision classification.  

\begin{table*}
\small
\begin{center}
\begin{tabular} {p{0.48\linewidth}|p{0.48\linewidth}}
\hline 
\multicolumn{2}{l}{\textbf{Draft 1}}\\
\multicolumn{2}{p{0.96\linewidth}}{
\textbf{(D1-1)} Tone has a lot to say for Louv. \textbf{(D1-2)} On account that Louv uses words to sound completely annoyed and disgusted with how far people have drifted, says he is very disgusted and annoyed. \textbf{(D1-3)} The beginning paragraph tells that scientists can now, at will, change the colors of butterfly wings. \textbf{(D1-4)} Telling how humans are in control, at will, with nature.} 
\\
\hline
\multicolumn{2}{l}{\textbf{Draft 2}}\\
\multicolumn{2}{p{0.96\linewidth}}{\textbf{(D2-1)} The way Louv “talks” throughout the essay is his tone. \textbf{(D2-2)} Using words to sound very annoyed and completely disgusted. \textbf{(D2-3)} In the beginning of the excerpt, Louv tells of what scientists are doing now with nature, such as changing the colors of butterfly wings. \textbf{(D2-4)} Telling how humans are in control, at will, with nature.}\\
\hline
\textbf{Gold-standard revision extraction} & \textbf{Automatic revision extraction}\\

(D1-1, D2-1), (D1-2, D2-2), (D1-3, D2-3), (D1-4, D2-4)
&
(D1-1, D2-1), \st{(D1-2, Null), (Null, D2-2), (D1-3, Null), (Null, D2-3)}, (D1-4, D2-4)\\

\textbf{Gold-standard revision classification} & \textbf{Automatic revision classification}\\
(D1-1, D2-1, Modify, Surface) & (D1-1, D2-1, Modify, Surface)\\
(D1-2, D2-2, Modify, Surface) & \st{(D1-2, Null, Delete, Reasoning)} \\
 & \st{(Null, D2-2, Add, Reasoning)} \\
(D1-3, D2-3, Modify, Surface) & \st{(D1-3, Null, Delete, Reasoning)} \\
 & \st{(Null, D2-3, Add, Reasoning)} \\
(D1-4, D2-4, Nochange) & (D1-4, D2-4, Nochange) \\
\hline
\end{tabular}
\end{center}
\caption{An example of pipeline revision identification errors (\st{striked}). A revision is represented as (D1-SentenceIndex, D2-SentenceIndex, RevisionOp, RevisionType) (e.g. (D1-1, D2-1, Modify, Surface)). In the example 6 revisions are identified. The revision extraction step aligns D1-2 and D1-3 wrongly as the syntactic similarities between the gold-standard sentences are not strong enough. The errors of the alignment step propagates to 4 false ``Reasoning'' revisions in the revision classification step.}
\label{table:errorexample}
\end{table*}

\section{Related Work}
The idea of using sequence labeling for revision identification derives from the work in \cite{zhang-litman:2016:N16-1}, where they focused on the revision classification step with the types of revisions used as labels. Revisions are transformed to a sequence of labels according to the gold-standard alignment information. In our work, the sentence alignment step is also included as a target of our identification\footnote{As the models in this paper are trained at the paragraph level, we assume the paragraphs were aligned and leave the discussion of paragraph alignment to the future work.}. We extend their work by grouping sentence alignment and revision type together into one label for joint identification.

As our tasks involve alignment, the problem in this paper can look similar to a labeled alignment problem, which can be solved with approaches such as CRFs \cite{blunsom2006discriminative} or structured perceptrons/SVMs \cite{moore2006improved}. For example, Blunsom and Cohn \shortcite{blunsom2006discriminative} utilized CRFs to induce word alignment between bilingual sentence pairs. In their work, each sentence in the source document is treated as a sequence. Sequence labeling is conducted on the source sentence and the index of the aligned word in the target sentence is used as the label. Features such as translation scores between words are used and the Viterbi algorithm is used to find the maximum posterior probability alignment for test sentences. Our problem is more complicated as our labels cover both the alignment and the revision type information. In labeled alignment, labels are used to represent the alignment information itself in \textbf{one} sequence. In revision identification, labels are used to represent the \textbf{interaction between two sequences} (the difference between sentences). Thus, our work utilized the revision operation (add/delete/modify) instead of the sentence index to mark the alignment information. Such design allows us to have the location information better coupled with the revision type information, and meanwhile allows us to update the alignment prediction by simply mutating the revision operation part of the labels.

The idea of sequence mutation is introduced from genetic  algorithms to generate possible sentence alignment corrections. There are works on tagging problems \cite{araujo2002part,alba2006natural,silva2013new} where genetic algorithms are applied to learn a best labeling or rules for labeling. However, our approach does not follow the standard genetic algorithm in that we do not have crossover operations and we stop mutating when the current generation is worse than last. The idea behind our seed generation approach is similar to Sequential Monte-Carlo (Particle-filter) \cite{khan2004mcmc}, where the sequence samples are generated by sampling labels according to their previous labels. In the paper we utilize a Long Short Term Memory (LSTM) \cite{hochreiter1997long} RNN to generate sample sequences as seeds. The advantage of  LSTM is that it can utilize long distance label information instead of just the label before.

\section{Joint Revision Identification}
\subsection{Problem and Approach Description}
As demonstrated in Table \ref{table:errorexample}, our task aims at the identification of the author's modifications from one draft to the other draft. Given the sentences from two drafts as the input, the system provides output in the format as (D1-1, D2-1, Modify, Surface), where the sentence alignment is used to record the revision locations and the revision type is used to record the author's revision purpose.  

\begin{figure*}
\centering
\includegraphics[scale=0.48]{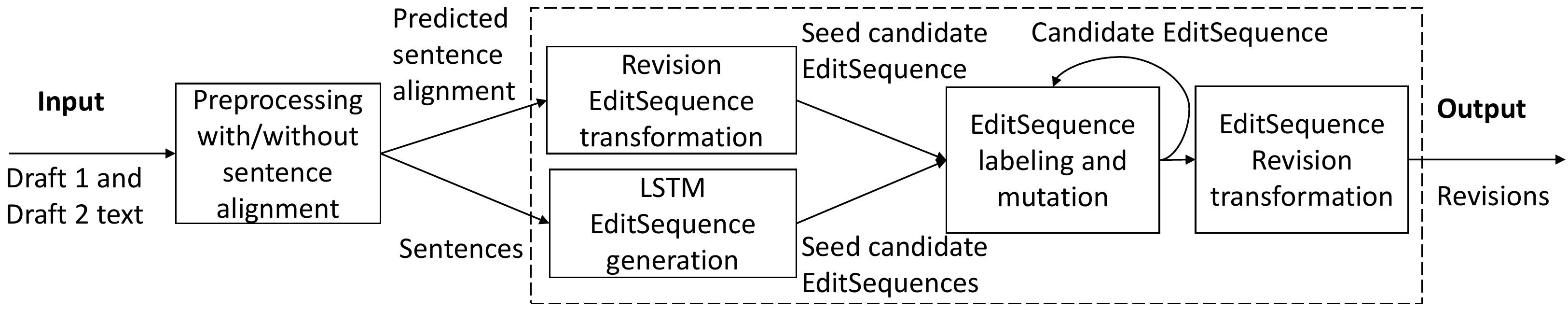}
\caption{Overall approach architecture. Components within the dashed box are covered in this paper. Notice that sentence alignment in the preprocessing step can be skipped with LSTM sequence generation.}
\label{figure:architecture}
\end{figure*}

Figure \ref{figure:architecture} demonstrates the workflow of our approach. The sentence alignment approach in \cite{zhang-litman:2014:W14-18} is first utilized to segment the essays into sentences and generate a sentence alignment prediction. Afterwards seed EditSequences are generated either using a LSTM network or by transforming directly from the predicted sentence alignment. The seed EditSequences are then labeled by the trained sequence labeling model. The candidate EditSequences are mutated according to the output of the sequence model. Finally the best EditSequence is chosen and transformed to the list of revisions. 

\subsection{Transformation between Revision and EditSequence Representation}
Instead of using the sentence indices as the alignment information as in other works \cite{blunsom2006discriminative}, this paper proposes \textbf{EditSequence} as a sequence representation of revisions. It incorporates both the alignment information and the revision type information in one sequence\footnote{Following ~\cite{zhang-litman:2016:N16-1}, we treat a revision that reorders two sentences as a Delete and an Add revisions.}. 

\textbf{EditStep} is defined as the basic unit of an \textbf{EditSequence}. An EditSequence contains a consecutive sequence of EditSteps. An EditStep unit contains 3 elements $(Op1, Op2, RevType)$. For a pair of revised essays (Draft1, Draft2), a cursor is created for each draft separately and we define $D1Pos, D2Pos$ to record cursor locations. $Op1$ and $Op2$ record the \textbf{action}s of the cursors. There are two cursor actions: \textbf{Move} (M) and \textbf{Keep} (K). \textbf{Move} indicates that the corresponding cursor is going to move to the position of the next sentence while \textbf{Keep} indicates that the cursor remains at the same location. $RevType$ records the revision type information. In this paper we follow \cite{zhang-litman:2016:N16-1}, where revision types include five types\footnote{\textit{Claim/Ideas (Claim)}, \textit{Warrant/Reasoning/Backing (Reasoning)}, \textit{Evidence}, \textit{General Content (General)} and \textit{Surface}} for sentences changed\footnote{Added/Deleted/Modified} and one type \textit{Nochange} when aligned sentences are identical. 
\begin{figure*}[tp]
\centering
\subfigure[M-M-Nochange]{\includegraphics[width=0.25\linewidth]{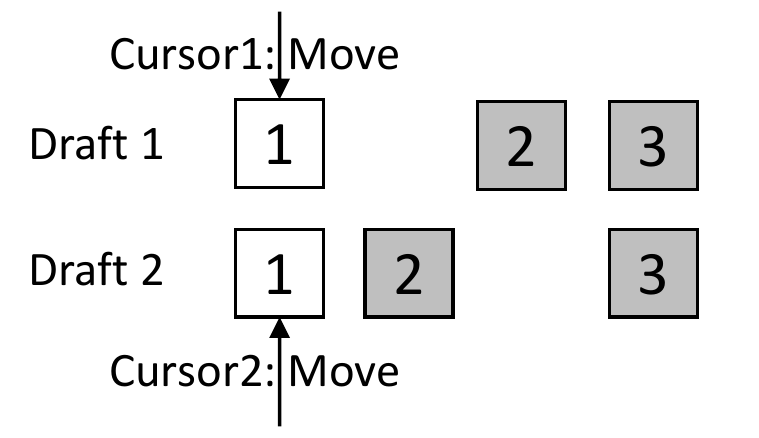}\label{fig:step1}}
\subfigure[K-M-Reasoning]{\includegraphics[width=0.19\linewidth]{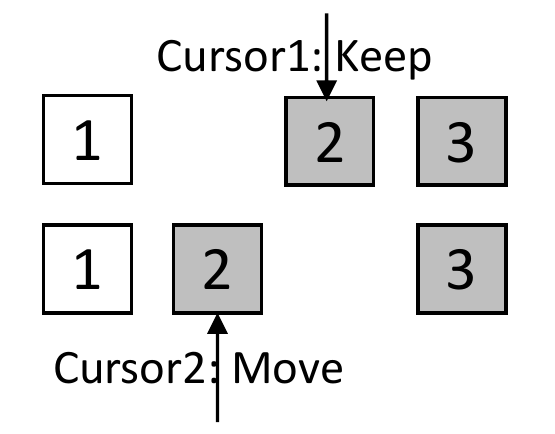}\label{fig:step2}}
\subfigure[M-K-Reasoning]{\includegraphics[width=0.20\linewidth]{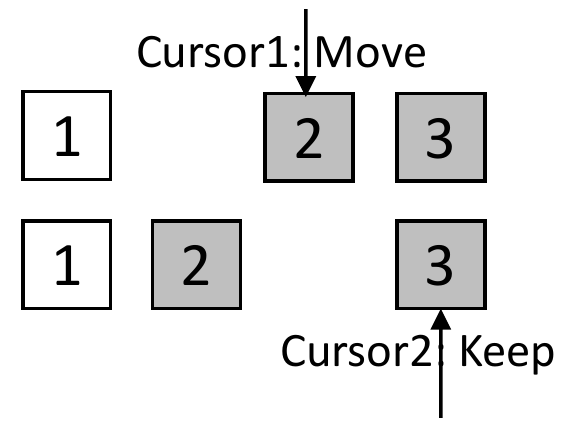}\label{fig:step3}}
\subfigure[M-M-Surface]{\includegraphics[width=0.22\linewidth]{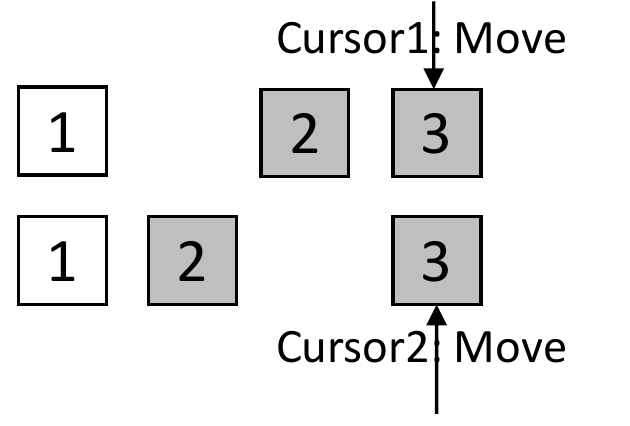}\label{fig:step4}}
\caption{Example of EditSequence transformation. The first row represents the sentences of the original essay (Draft1) and the second row represents the sentences of the revised essay (Draft2). The vertical direction indicates sentence alignment. The shadowed sentences are revised and there are three revisions: (Null, 2, Add,  Reasoning), (2, Null, Delete,  Reasoning) and (3, 3, Modify, Surface).  With the cursors, we transform the revisions to 4 consecutive EditSteps from left to right and generate the sequence representation (M-M-Nochange -$>$ K-M-Reasoning -$>$ M-K-Reasoning -$>$ M-M-Surface). } 
\label{fig:editsequence}
\end{figure*}

\textbf{Revisions to EditSequence}. Figure \ref{fig:editsequence} demonstrates how we transform from the revision representation used in prior works to our sequential representation EditSequence. In Figure \ref{fig:step1}, the cursors of the two drafts start at the beginning of the segment with $D1Pos$ and $D2Pos$ set to 1. Given that sentence 1 in Draft1 is the same as sentence 1 in Draft 2, both cursors move to the next sentence and we generate an EditStep (M, M, Nochange). In Figure \ref{fig:step2}, $D1Pos$ and $D2Pos$ are set to 2 according to the action of the previous step. In the example, sentence 2 in Draft 2 is an added \textit{Reasoning} sentence, thus we generate a new EditStep (K-M-Reasoning) by keeping the cursor of Draft 1 in its current position (for comparison at the next step) and moving the cursor of Draft 2. Similarly, we move the cursor of Draft 1 in Figure \ref{fig:step3}. In Figure \ref{fig:step4}, $D1Pos$ and $D2Pos$ are set to 3. Sentences at the position are aligned to each other and both cursors are thus moved. Each \textbf{EditStep} is assigned a label as $Op1$-$Op2$-$RevType$ and thus we generate a labeled sequence representation of revisions. As there are only three possible Op combinations (M-M, K-M, M-K)\footnote{At least one of the cursors has to move.}, the total number of possible labels is $3 \times RevisionClassNum$. 

\textbf{EditSequence to Revisions}. The sequence transformation step is reversible and we can infer all the revisions according to the sequence of edits. Head of the EditStep label indicates the revision location: a label starting with ``M-M'' indicates that two sentences are aligned, ``M-K'' indicates that a sentence is deleted while ``K-M'' indicates that a sentence is added. Tail of the label corresponds to the revision type.

\subsection{EditSequence Labeling and EditSequence Mutation}
\textbf{For our first hypothesis}, we conduct sequence labeling on EditSequence and use $RevType$ of the labeled sequence as the results of revision classification. \textbf{For our second hypothesis}, we utilize both the likelihood provided by the sequence labeler and the ($Op1$,$Op2$) information of labels to correct sentence alignments.  

\paragraph{EditSequence Labeling}
Given a candidate EditSequence, sequence labeling is conducted to assign labels to each EditStep in the sequence. The $RevType$ part of the assigned label is used as the revision type. Conditional Random Fields  (CRFs) \cite{lafferty2001conditional} is utilized for labeling\footnote{CRFSuite \cite{CRFsuite} is used in implementation.}. Features used in \cite{zhang-litman:2015:bea} are reused, which include unigrams and three feature groups.

\textit{Location group.} For each EditStep, we record its corresponding $D1Pos$ and $D2Pos$ as features, We also record whether the $D1Pos$ and $D2Pos$ are at the beginning/end of the paragraph/essay.

\textit{Textual group.} For each EditStep, we extract features for the aligned sentences pair ($D1Pos$, $D2Pos$). Features include sentence length (in both drafts), edit distance between aligned sentences and the difference in sentence length and punctuation. We not only calculate the edit distance between sentence pair ($D1Pos$, $D2Pos$) but also for pairs ($D1Pos$, $D2Pos$+1)\footnote{If D2Pos+1 does not exceed paragraph boundary} and ($D1Pos$+1, $D2Pos$).   

\textit{Language group.} Part of speech (POS) unigrams and difference in POS counts are encoded. Again features are extracted for pairs ($D1Pos$, $D2Pos$+1) and ($D1Pos$+1, $D2Pos$) besides ($D1Pos$, $D2Pos$).   

\paragraph{EditSequence mutation}
Besides assigning labels to the sequence, the CRFs model also provides us the likelihood of each label and the likelihood of the whole sequence. We compare the likelihood between sequences to decide which sequence is a better labeling. Within one sequence, we compare the likelihood between EditSteps to decide which EditStep is most likely to be corrected. Besides using the likelihood of each EditStep, we also compare the ($Op1$, $Op2$) information with the ($Op1$, $Op2$) information of the prior candidate EditSequence. We call it \textbf{collision} when such information does not match, which indicates that the candidate's alignment does not follow a typical sequence pattern and suggests correction. 

\begin{figure*}
\centering
\includegraphics[scale=0.45]{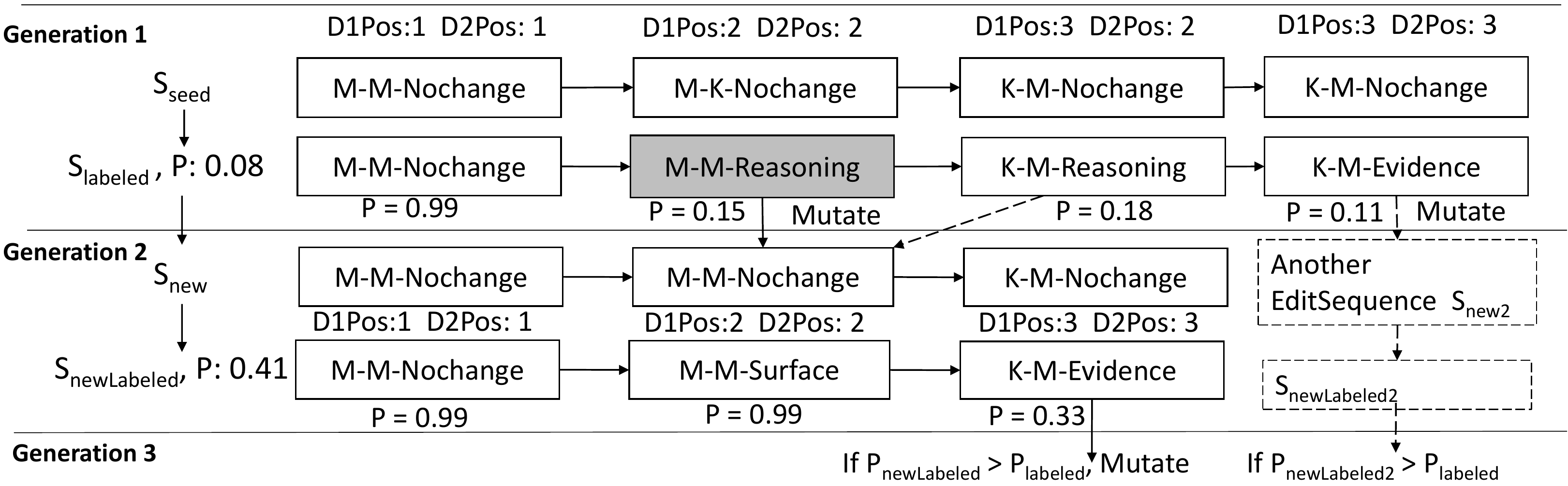}
\caption{Example of EditSequence update. Two EditSequences can be mutated from $S_{labeled}$: one from the EditStep with collision (the shadowed EditStep) and one from the EditStep with the lowest likelihood (the last EditStep). The first generation (seed sequences) will always be mutated, while the other generations will only mutate if they have a larger likelihood than the prior generation. Note that only $RevType$ in labeled sequences ($S_{labeled}$ or $S_{newLabeled}$)  will be used as the type of revisions.}
\label{figure:mutation}
\end{figure*}

We borrow the idea of ``mutation'' from genetic algorithms to generate possible corrections of sentence alignment. There are three possible kinds of ``mutation'' operations. \textbf{(1) ``M-M'' to ``M-K'' or ``M-M'' to ``K-M''}. This indicates breaking an alignment of sentences to one \textit{Delete} revision and one \textit{Add} revision. Thus for a EditStep with tag ``M-M-Type'', we would remove the step from the sequence and add two new steps ``M-K-Nochange'' and ``K-M-Nochange''. Notice that here \textit{Nochange} is a dummy label and will be replaced in the next round of labeling. \textbf{(2) ``M-K'' to ``M-M'' or ``K-M'' to ``M-M''}. This indicates aligning a deleted/added sentence to another sentence. Depending on the labeling of the following EditStep, the operation can be different. ``M-K'' followed by ``K-M''\footnote{Or ``K-M'' is followed by ``M-K''.} indicates that the aligned sentence in Draft 2 is not aligned to other sentences. For example in Figure \ref{figure:mutation}, the second EditStep (M-K-Nochange) is followed by EditStep (K-M-Nochange), which indicates that Sentence 2 (Draft 2) has not been aligned to other sentences and aligning sentence 2 (Draft 1) will not impact the alignment of Sentence 2 (Draft 2). In that case, we remove the two steps and add a step ``M-M-Nochange''. ``M-K'' followed by ``M-M'' indicates that the aligned sentence has been aligned to other sentences. For that cases, we need to remove the ``M-K'' and ``M-M'' step and add two steps ``M-M-Nochange'' and ``M-K-Nochange'' for the misaligned sentence. \textbf{(3) ``M-K'' to ``K-M'' or ``K-M'' to ``M-K''}. This means changing from \textit{Delete} to \textit{Add}. This is similar to the previous case, where the mutation operation depends on the labeling of the following EditStep. If the following EditStep starts with ``M-M'', it indicates that the sentence in the \textit{Add} revision is aligned and we need to break the existing alignments and add a ``M-K'' EditStep besides changing ``M-K'' to ``K-M''. 

 Figure \ref{figure:mutation} provides an example of the EditSequence update process. The process starts with seed candidate sequences as the first generation, the first generation will always be mutated. For a seed EditSequence $S_{seed}$ and its labeled sequence $S_{labeled}$, the alignment part of their EditStep labels are compared to check for collision. For every collision detected, we mutate $S_{seed}$ to generate one new candidate sequence $S_{new}$ as a member of the next generation. After the mutation of the first generation is complete, all $S_{new}$ in the new generation are labeled with CRFs again. The new labels provide us new revision types within the new alignments.  If the likelihood of the labeled sequence $S_{newLabeled}$ is larger than $S_{labeled}$, it indicates that the sentence alignment in $S_{new}$ is more trustworthy than the alignment in $S_{seed}$, thus $S_{new}$ should be further mutated to see if the alignment can be further improved. Otherwise we do not further mutate $S_{new}$. We keep mutating the EditSequences until we cannot conduct any further mutation.  For the labeled EditSequences in all generations, we first select sequences with minimum number of collisions and then select the sequence with the maximum sequence likelihood. The ($Op1$, $Op2$) of labels are used as results of revision extraction and $RevTypes$ are used for revision classification. Through the process, sequence labeling provides likelihood for both alignments and revision types, while sequence mutation provides new possible sequences for labeling.

\subsection{Seed Candidate EditSequence Generation}
\begin{figure}
\centering
\includegraphics[scale=0.3]{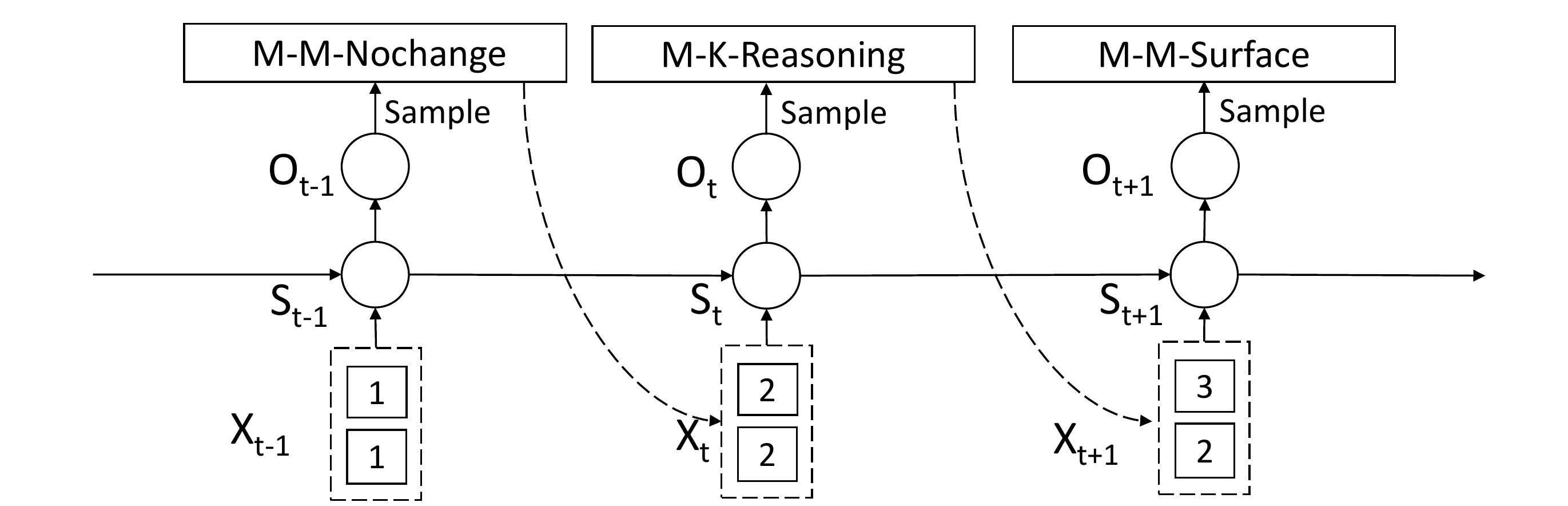}
\caption{LSTM recurrent neural network for generating candidate sequences. $X$ are features extracted according to the location of the cursors. For example, $X_{t-1}$ corresponds to features extracted when sentence index in Draft 1 is 1 and sentence index in Draft 2 is 1.}
\label{figure:lstm}
\end{figure}

For a paragraph with $m$ sentences in the first draft and $n$ sentences in the second draft, there is a total of $\binom{m+n}{n} = \frac{(m+n)!}{m!n!}$ possible sequences\footnote{With m sentences of Draft 1 set, there are m+n slots to put in the n sentences of Draft 2.}. While theoretically we can first generate a sequence without sentence alignment (all sentences in Draft 1 treated as deleted and all sentences in Draft 2 treated as added) as the seed sequence and keep mutating until the best sequence is found, such process is too computationally expensive and is likely to fall into local optima during mutation. Thus an approach is needed for the generation of high-likelihood seed EditSequences. We propose two approaches for sequence generation, one based on the revision extraction method proposed in \cite{zhang-litman:2014:W14-18}, the other based on automatic sequence generation with LSTM.

\begin{table*}
\small
\begin{center}
\begin{tabular}{c|c|c|c|c|c|c|c|c}
\hline
\textbf{Corpus} & \textbf{\#Sentence} & \textbf{\#Paragraph} & \textbf{\#Claim} & \textbf{\#Reasoning} & \textbf{\#Evidence} & \textbf{\#General} & \textbf{\#Surface} & \textbf{\#Nochange} \\
\hline
A  & 1776-$>$2495 & 285-$>$362 & 111 & 390 & 110 & 356 & 300 & 1265\\
B  & 1596-$>$1791& 295-$>$310& 76 & 327 & 34 & 216 & 391 & 917\\
\hline
\end{tabular}
\end{center}
\caption{Statistics and revision distribution in corpora A and B, 1776-$>$2495 indicates 1776 sentences in the first draft and 2495 sentences in the second draft. The numbers can be further summed up as 967 Content (111 Claim, 856 Support), 300 Surface and 1265 Nochange for A, 653 Content (76 Claim, 577 Support), 391 Surface and 917 Nochange for B}
\label{table:data}
\end{table*}

\begin{table*}
\small
\begin{tabular}{p{0.12\textwidth}|l|p{0.37\textwidth}|p{0.35\textwidth}}
\hline
\textbf{Groups} & \textbf{Model} & \textbf{Revision extraction (sentence alignment)} & \textbf{Revision classification} 
\\
\hline
Baseline (Base) & Pipeline & Based on sentence similarity \cite{zhang-litman:2014:W14-18} & CRF sequence labeling, using revision type as label \cite{zhang-litman:2016:N16-1}\\
\hline
1Best & Joint & \cite{zhang-litman:2014:W14-18} + EditSequence mutation& CRF sequence labeling, using both revision type and alignment as label \\
\hline
+NCandidate (+NC) & Joint & \cite{zhang-litman:2014:W14-18} + LSTM EditSequence generation + EditSequence mutation &  CRF sequence labeling, using both revision type and alignment as label\\ 
\hline
\end{tabular}
\caption{Description of three implemented approaches}
\label{table:featuregroups}
\end{table*}

\textbf{1-Best EditSequence generation based on alignment prediction} During preprocessing, the essays are segmented into sentences and sentences are aligned following \cite{zhang-litman:2014:W14-18}. A logistic regression classifier is first trained on the training data with Levenshtein distance as the feature and alignment is conducted using Nelken's global alignment approach \cite{nelken2006towards} based on the likelihood provided by the classifier. As the number of essays in the dataset is limited, we construct sequences at the paragraph level. We trained our models on paired paragraphs assuming paragraphs have been aligned. For each paragraph pair, an EditSequence is generated following the sequence transformation method with $RevType$ of all EditSteps set to \textit{Nochange}\footnote{$Nochange$ is a just a placeholder as the real $RevType$ are to be labeled in the labeling step.}.

\textbf{N-Candidate EditSequence generation with LSTM network} The 1-best approach can provide a good sequence to start with, however, it is more likely to fall into local optima in the labeling step with only one seed candidate. Thus we also trained LSTM to generate multiple possible candidates. As demonstrated in Figure \ref{figure:lstm}, we construct the neural network with LSTM units. Due to the size limit of our current training data, we only include one layer of LSTM units to reduce the number of parameters in the network. Each EditStep is treated as a time step in the neural network. According to the $D1Pos$ and $D2Pos$ property of the EditStep, we extract features $X$ as the input to the neural network. The same set of features used in the sequence labeling step is used. The model transforms the input to hidden state $S$, where hidden state $S_{t-1}$ at time (t-1) is used together with input $X_{t}$ to predict the hidden state $S_{t}$ at time t. A softmax layer is added on the top of the hidden state to predict $O_{t}$, which describes the probability distribution of the sequence labels. At the training step, we fit the model with EditSequences transformed from revisions between the paragraphs. At the generation step we start with both $D1Pos$ and $D2Pos$ set to 1 and extract features for $X_1$. In each time step, a label is sampled according to the probability distribution $O_t$.  According to the sampled label, we change the positions of $D1Pos$ and $D2Pos$ to extract the features for $X_2$. In the example, the sampled label at $X_{t-1}$ is M-M-Nochange, this label moves $D1Pos$ and $D2Pos$ to (2,2) and the $X_t$ is extracted and used together with $S_{t-1}$ to predict $S_{t}$. According to the probability distribution $O_{t}$, a new label is sampled and the result is used to move the cursors for the next EditStep. The process is repeated until an EditSequence is generated for the whole paragraph pair. We repeat the algorithm until N candidates are collected.

\section{Experiments and Results}
\paragraph{Data}
As in Table \ref{table:data}, our experiments use the data used in \cite{zhang-litman:2016:N16-1}, which consists of Drafts 1 and 2 of papers written by high school students taking AP-English courses; papers were revised after receiving and generating peer feedback. Corpus A contains 47 paper draft pairs about placing contemporaries in Dante's Inferno. Corpus B contains 63 paper draft pairs explaining the rhetorical strategies used by the speaker/author of a previously read lecture/essay. Both corpora were double coded (Kappa for A: 0.75, B: 0.69) and gold standard labels were created upon agreement.

\paragraph{Experiments}

We conducted experiments using different revision type settings. In \cite{zhang-litman:2015:bea}, the annotated \textit{Claim}, \textit{Reasoning}, \textit{Evidence} and \textit{General Content} were grouped together as one \textit{Content} revision category\footnote{In contrast to the \textit{Surface} revision type, \textit{Content} represent revisions that change the information of the essay.}. In our work we in addition group the last three categories together as one \textit{Support} category\footnote{Content revisions that support the claim of the essay.}. We first evaluate the performance of sentence alignment and \textit{Content} vs. \textit{Surface} vs. \textit{Nochange} revision classification (\textbf{3-class}). Then we experimented with \textit{Claim} vs. \textit{Support} vs. \textit{Surface} vs. \textit{Nochange} (\textbf{4-class}). Finally we used all revision categories (\textbf{6-class}). For each experiment, three approaches are compared as in Table \ref{table:featuregroups}: \textbf{Baseline}, \textbf{1Best} and \textbf{+NCandidate}. 10 draft pairs from Corpus B were used as the development set for setting up parameters of LSTM\footnote{LSTM implemented with deeplearning4j (\url{http://deeplearning4j.org}) with epoch set to 1, iteration numbers to 100 and output dimension of the first layer to 100} and choosing N. N is set to 10 for all our experiments. Afterwards 10-fold (student) cross-validation were conducted on both corpora A and B. The same set of data folds and features were used for all three approaches. The training folds in each round will be used for training both CRFs and LSTM. For evaluation we used alignment accuracy\footnote{$\frac{2 \times AgreedAlignment}{\#Draft1Sentences + \#Draft2Sentences} $, adapted from \cite{zhang-litman:2014:W14-18}.} to measure the accuracy of \textbf{revision extraction} and precision/recall to measure the result of \textbf{revision identification}. Precision is calculated as $\frac{\#Correct Revisions}{\#Predicted Revisions}$ and Recall is calculated as $\frac{\#Correct Revisions}{\#GoldStandardRevisions}$.

\begin{table}
\small
\begin{center}
\begin{tabular}{p{0.10\linewidth}|l|l|l|l|l}
\hline
&  &  & \textbf{Extraction} & \multicolumn{2}{l}{\textbf{Classification}} \\
\cline{4-6}
& & & \textbf{Accuracy} &\textbf{Prec} & \textbf{Recall} \\
\hline
\multirow{6}{*}{\parbox{\linewidth}{\textbf{3-class} \\
}} &
A & Base & 0.940 & 0.780 & 0.830 \\
 &  & 1Best & 0.948$\ast$ & 0.801$\ast$ & 0.859$\ast$ \\
 &  & +NC & \textbf{0.957}$\ast\ddagger$ & \textbf{0.815}$\ast\ddagger$ & \textbf{0.875}$\ast\ddagger$ \\
   \cline{2-6}
& B & Base & 0.928 & 0.780  & 0.834\\
 &  & 1Best & 0.930  & 0.782 & 0.840\\
 &  & +NC & \textbf{0.934} & \textbf{0.788} & \textbf{0.848}$\ddagger$\\
   \hline
  \multirow{6}{*}{\parbox{\linewidth}{\textbf{4-class}\\ 
  }}
  &
A & Base & \textbf{0.940} & 0.647 & 0.685 \\
&   & 1Best & 0.937  & 0.648  & 0.703$\ast$ \\
&   & +NC & \textbf{0.940} & \textbf{0.652} & \textbf{0.723}$\ast\ddagger$ \\
   \cline{2-6}
& B & Base& 0.928 & 0.595  & 0.627\\
&   & 1Best & 0.935$\ast$  & 0.620$\ast$ & 0.654$\ast$\\
&   & +NC & \textbf{0.944}$\ast\ddagger$ & \textbf{0.647}$\ast\ddagger$ & \textbf{0.702}$\ast\ddagger$\\
   \hline
   \multirow{6}{*}{\parbox{\linewidth}{\textbf{6-class}\\
   }} &
A & Base  & 0.940 & 0.397 & 0.376 \\
 &  & 1Best & 0.940  & 0.411$\ast$  & 0.390$\ast$ \\
 &  & +NC  & \textbf{0.948}$\ast$ & \textbf{0.427}$\ast\ddagger$ & \textbf{0.406}$\ast\ddagger$ \\
   \cline{2-6}
& B & Base  & 0.928 & \textbf{0.400}  & \textbf{0.344}\\
&   & 1Best & 0.930  & 0.393 & 0.339\\
&   & +NC  & \textbf{0.936} & 0.390 & 0.338\\
   \hline
\end{tabular}
\end{center}
\caption{The average of 10-fold (student) cross-validation results on Corpora A and B. Alignment accuracy, Unweighted average precision/recall are reported.  $\ast$ indicates significantly better than the baseline, $\ddagger$ indicates significantly better than 1Best (Paired T-test, p $<$ 0.05), \textbf{Bold} indicates best result  }
\label{table:results}
\end{table}

\paragraph{Results}
Table \ref{table:results} demonstrates our experimental results. We first compare the pipeline baseline with our joint model using 1Best seed EditSequence. With 3 revision types (3-class), the joint model achieves significantly better performance than the baseline on Corpus A for both revision extraction (sentence alignment) and revision classification. It also shows better performance on Corpus B (while not significant). The improvement on the precision/recall of revision classification supports our first hypothesis that alignment information can improve the accuracy of revision classification. The improvement on sentence alignment supports our second hypothesis that the patterns of predicted revisions can be used to correct the false alignments. We notice that the number of revision types impacts the performance of the model. On corpus A, the model shows significantly better performance than the baseline in almost all experiments. While on corpus B, the model yields significantly better performance in 4-class experiment. The impact of revision types on our model can be two-fold. On the one hand, more revision types indicates more detailed sequence information, which improves the chance of recognizing problems in sentence alignment. On the other hand, the increase of revision types increases the difficulty of sequence labeling, which in return can hurt the performance of joint identification. We leave the error analysis of performance difference between different revision types to the future work.  

Next, we compare results using \textbf{1Best} and \textbf{+NCandidate} EditSequences. We observe that using generated sequences improves the 1Best performance, yielding the best result on almost all experiments (except on Corpus B with 6 revision types).  We counted the number of generations in EditSequence mutation for both 1Best and +NCandidate on 3-class experiment. Results show that the 1Best approach will stop mutating after an average of 1.2 generations while +NCandidate stops mutating after an average of 2.3. This suggests that our approach prevents the model from easily falling into local optima.

\section{Conclusion and Future Work}
In this paper we propose a joint identification approach for argumentative writing revisions. For the two different sub tasks of revision identification (revision location extraction and revision type classification), we transform the location representation to a revision operation format and incorporate it together with the revision type into one label. The two different tasks are thus transformed to one joint sequence labeling task. With this design, the likelihood of a labeled sequence indicates not only the likelihood of sentence alignments but also the likelihood of the revision types. We utilize the mutation idea from genetic algorithms to iteratively update the labeling of sequences. LSTM is utilized to generate seed candidate EditSequences for mutation. Results demonstrate that our approach improves the performance of both tasks.

In this paper the effect of neural networks is limited by the data size. For the future work we would like to explore our approach on other larger writing revision datasets \cite{leecityu} to fully take advantage of LSTM. Another problem that has not been addressed in this paper is that the paragraphs are assumed to be aligned. To fully automatize our model, we plan to construct an accurate automatic paragraph alignment model \cite{barzilay2003sentence} based on topic information \cite{blei2003latent} as the preprocessing step. 





\bibliography{eacl2017}
\bibliographystyle{eacl2017}

\end{document}